\documentclass[letterpaper, 10 pt, conference]{ieeeconf}  

\IEEEoverridecommandlockouts                              

\usepackage{url}
\usepackage{caption}
\usepackage{subcaption}
\usepackage{graphicx}
\usepackage{cite}
\usepackage{multirow}
\usepackage{enumerate}
\usepackage{amsmath}
\usepackage{etoolbox}
\usepackage{placeins}
\usepackage{hyperref}
\usepackage{setspace} 
\usepackage{romannum}
\usepackage{wasysym}

\usepackage{bbding}
\usepackage{pifont}
\usepackage{amssymb}
\usepackage{xcolor}
\usepackage[outline]{contour}
\DeclareMathAlphabet\mathbfcal{OMS}{cmsy}{b}{n}

\title{\LARGE \bf GraspSAM: When Segment Anything Model Meets Grasp Detection}

\author{Sangjun Noh, Jongwon Kim, Dongwoo Nam, Seunghyeok Back, Raeyoung Kang, Kyoobin Lee†
\thanks{All authors are with the School of Integrated Technology (SIT), Gwangju Institute of Science and Technology (GIST), Cheomdan-gwagiro 123, Buk-gu, Gwangju 61005, Republic of Korea. 
† Corresponding author: Kyoobin Lee {\tt\small kyoobinlee@gist.ac.kr}}%
}

\begin{document}

\maketitle
\thispagestyle{empty}
\pagestyle{empty}

\begin{abstract}
Grasp detection requires flexibility to handle objects of various shapes without relying on prior object knowledge, while also offering intuitive, user-guided control. In this paper, we introduce GraspSAM, an innovative extension of the Segment Anything Model (SAM) designed for prompt-driven and category-agnostic grasp detection. Unlike previous methods, which are often limited by small-scale training data, GraspSAM leverages SAM’s large-scale training and prompt-based segmentation capabilities to efficiently support both target-object and category-agnostic grasping. By utilizing adapters, learnable token embeddings, and a lightweight modified decoder, GraspSAM requires minimal fine-tuning to integrate object segmentation and grasp prediction into a unified framework. Our model achieves state-of-the-art (SOTA) performance across multiple datasets, including Jacquard, Grasp-Anything, and Grasp-Anything++. Extensive experiments demonstrate GraspSAM’s flexibility in handling different types of prompts (such as points, boxes, and language), highlighting its robustness and effectiveness in real-world robotic applications. Robot demonstrations, additional results, and code can be found at \href{https://gistailab.github.io/graspsam/}{https://gistailab.github.io/GraspSAM/}.
\end{abstract}


\section{INTRODUCTION}
As robots become more prevalent in household and industrial environments, their ability to perform efficient object manipulation is increasingly important. Prompt-based grasping techniques have emerged as a promising approach for enabling robots to quickly respond to user instructions, such as GUI clicks, eye-gazing, or text-based prompts. These techniques are crucial in applications like collaborative manufacturing, warehouse automation, and assistive care, where handling a wide variety of objects is essential. However, despite advancements in deep learning and grasp detection models \cite{mahler2017dex, morrison2018closing, kumra2020antipodal, ainetter2021end}, many existing methods are limited by small-scale training data, rely on separate networks for object identification and grasp prediction, and cannot directly handle prompt-based inputs. This restricts their adaptability and scalability in real-world scenarios, where category-agnostic and user-guided grasping is needed.


\begin{figure}[ht]
    \centering
        \begin{subfigure}[t]{\columnwidth}
            \centering
            \includegraphics[width=\textwidth]{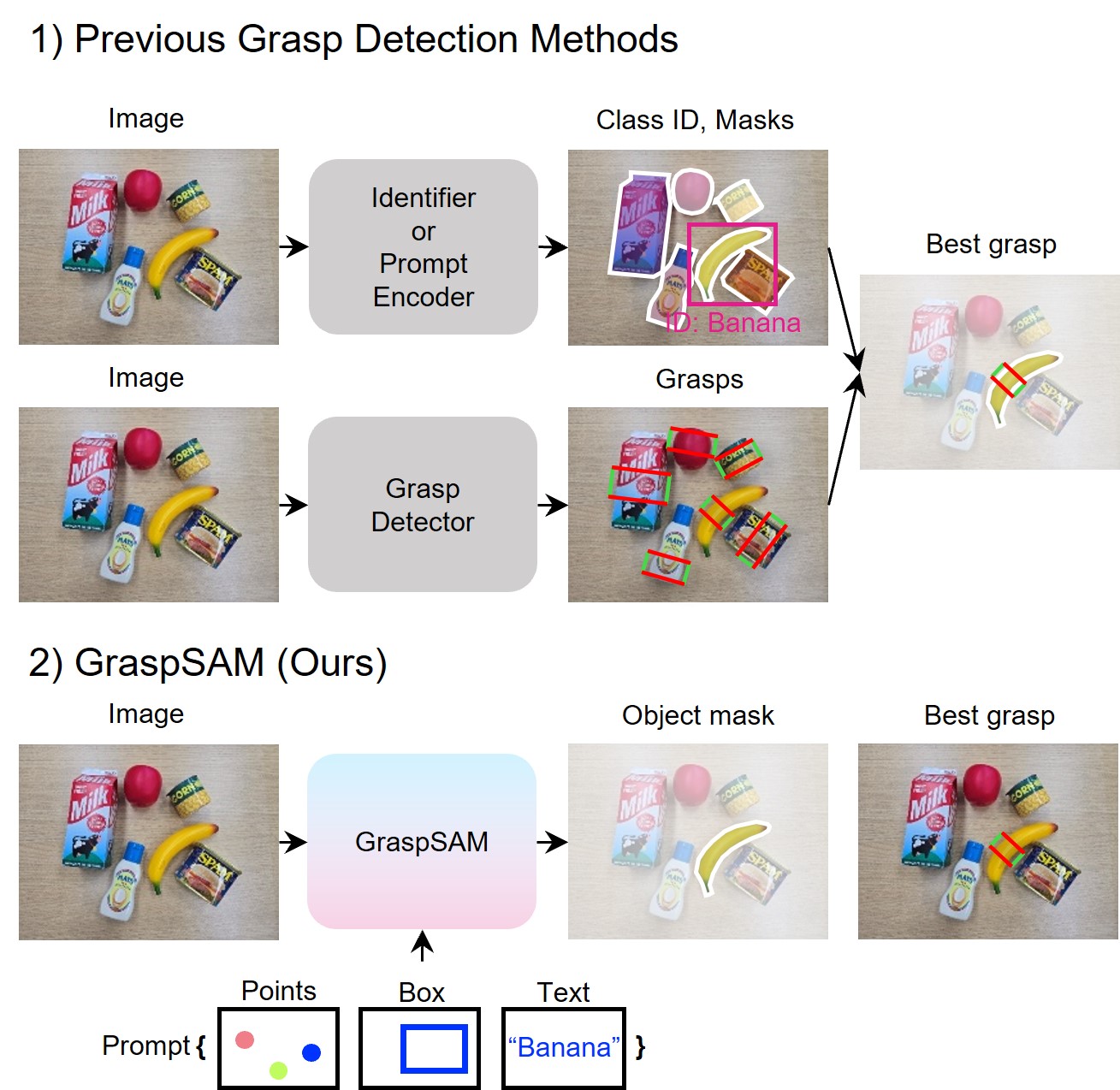}
        \end{subfigure}
        
    \caption{Conventional methods use separate networks for object identification and grasp prediction, while GraspSAM (Ours) predicts both the object mask and grasps from a single RGB image and prompt in a single step.
}    
    \label{fig:task_comparison}
\end{figure}

To overcome these challenges, we introduce GraspSAM, the first approach to extend the Segment Anything Model (SAM)\cite{kirillov2023segment} for end-to-end grasp detection. SAM’s powerful zero-shot segmentation capabilities and ability to generalize to diverse object types make it an ideal foundation for grasping tasks. However, adapting SAM for grasp detection presents unique challenges, particularly in combining object identification with grasp prediction in a seamless manner. We address this by proposing a minimal adaptation strategy that introduces lightweight token learning and a few additional parameters to SAM's decoder. This allows GraspSAM to unify object identification and grasp planning into a single process, reducing the computational complexity and eliminating the need for multiple networks.

We conducted extensive experiments to evaluate GraspSAM's ability to jointly learn object segmentation and grasp detection, while preserving SAM’s strengths. Notably, we extended the evaluation to include category-agnostic grasp detection, where GraspSAM achieved state-of-the-art (SOTA) performance. Grasp detection was tested using three prompt types (point, box, and language) across several benchmarks, including Jacquard\cite{depierre2018jacquard} , Grasp-Anything\cite{vuong2023grasp} , and Grasp-Anything++\cite{vuong2024language} , further demonstrating GraspSAM’s superior performance. Additionally, GraspSAM outperformed previous two-stage methods in real-world experiments, showcasing its practicality and effectiveness in diverse scenarios. These results confirm that SAM’s strong segmentation capabilities are well-suited for integration into grasp detection tasks, enabling more efficient and adaptable robotic manipulation. Our contributions are as follows: 
\begin{itemize} 
\item{We extend SAM for end-to-end grasp detection, enabling prompt-driven, category-agnostic object grasping without the need for separate networks for object identification and grasp prediction.} 
\item{We introduce lightweight token learning and small parameter additions to SAM's decoder, enabling efficient adaptation without extensive fine-tuning.} 
\item{We demonstrate SOTA performance on category agnostic grasp detection and prompt driven grasp detection for Jacquard, Grasp-Anything, and Grasp-Anything++ benchmarks.} 
\item{We validate GraspSAM’s real-world applicability by outperforming existing two-stage method in practical grasp experiments, show its effectiveness in diverse scenarios.} 
\end{itemize}

\section{Related Work}
\noindent\textbf{Grasp Detection.} Grasp detection is an essential for robotic manipulation, allowing robots to interact with their environment. Traditional methods\cite{murray2017mathematical, bicchi2000robotic} used geometric analysis to determine grasp points, but these approaches required 3D object models, limiting their effectiveness in real-world settings. With the advent of deep learning, grasp detection improved significantly, initially focusing on single-object grasping\cite{mahler2017dex} and later expanding to handle multiple objects\cite{morrison2018closing, kumra2020antipodal, ainetter2021end, wang2022transformer}. However, these deep learning models often required separate networks for target object identification(i.g., classification, segmentation). To address these limitations, our approach, GraspSAM, integrates object-specific prompts to detect grasp points directly, removing the need for separate identification networks. This unified method simplifies the grasp detection process and enhances performance in real-world applications, improving efficiency and adaptability for human-robot collaboration in various domains.

\noindent\textbf{SAM Families.} 
Segment Anything Model (SAM) generates object masks from prompts (e.g., points, bounding boxes, text), enabling zero-shot segmentation across diverse datasets. SAM's large image encoder and prompt-based decoder make it versatile for segmentation tasks, but its size and computational demands limit its use in real-time applications. To address these limitations, various SAM models\cite{zhao2023fast, songa2024sam}, include MobileSAM\cite{mobile_sam} and EfficientSAM\cite{xiong2024efficientsam} were proposed. MobileSAM reduces model size with a compact backbone, while EfficientSAM applies pruning and quantization to lower computational costs, making both more suitable for resource-constrained environments. While SAM performs well overall, its segmentation quality decreases when handling objects with intricate details or complex boundaries. HQ-SAM\cite{ke2024segment} improves precision by introducing a High-Quality Output Token in the mask decoder, leveraging early and final ViT features to enhance object detail. Unlike other SAM variants, we introduce an end-to-end framework, GraspSAM, which integrates object segmentation and grasp detection tasks.

\noindent\textbf{Fine-tuning for Foundation Models.} 
In recent years, adapter-based approaches\cite{hu2021lora, chen2022vitadapter, chen2022adaptformer, wei2024stronger} have gained significant attention as an efficient way to adapt large foundation models without full fine-tuning. By freezing most of the model's parameters and updating only small, task-specific layers (adapters), these methods reduce computational overhead while maintaining the model's core capabilities. Notable examples include LoRA (Low-Rank Adaptation)\cite{hu2021lora}, which has been applied to large language models, and SAM-adapter\cite{chen2023sam}, which fine-tunes SAM for medical image segmentation by inserting MLP layers between the encoder blocks. These techniques have proven effective in adapting foundation models to specialized tasks across various domains. GraspSAM utilizes an adapter-based method to optimize SAM for grasp detection, while preserving its powerful zero-shot segmentation capabilities.

\begin{figure*}[ht!]
    \centering
        \includegraphics[width=0.90\textwidth]{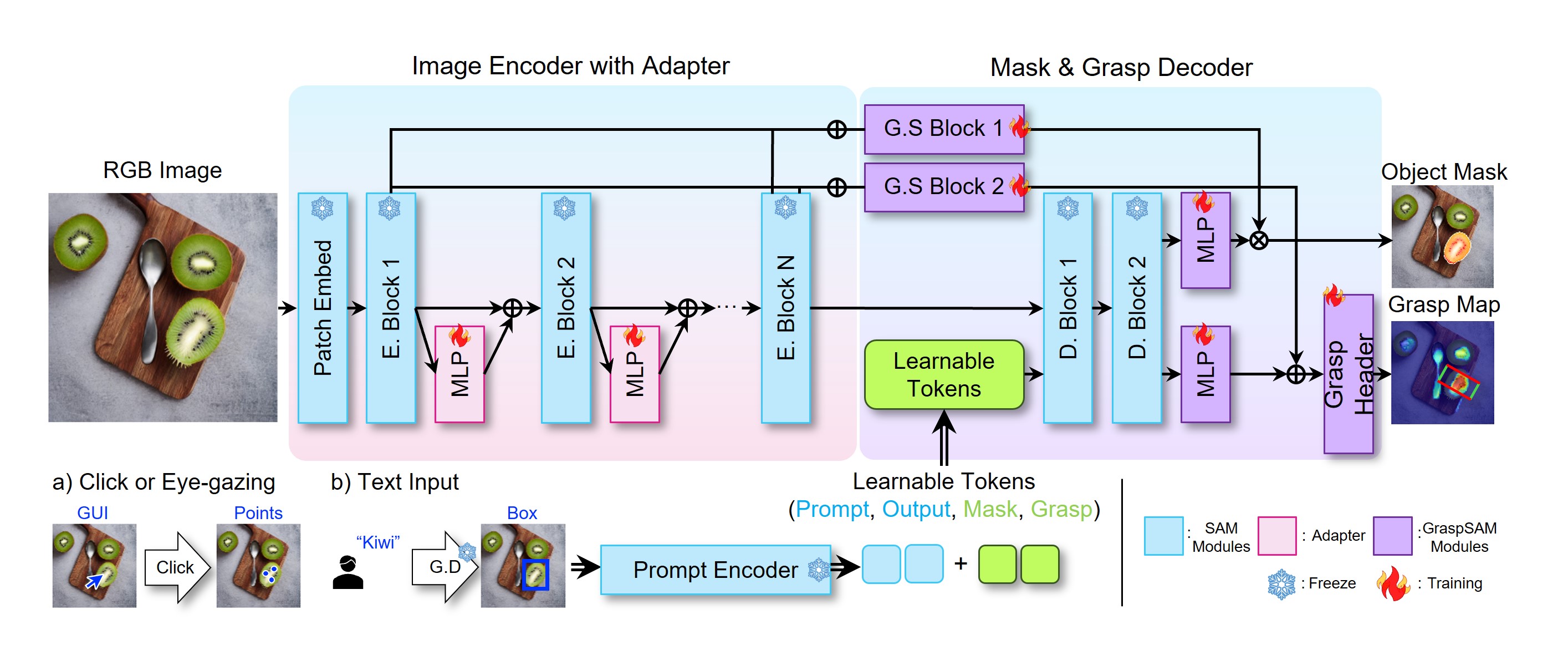}
    \caption{\textbf{GraspSAM overall pipeline.} GraspSAM builds upon the zero-shot capabilities of the SAM by adding an adapter for the image encoder, a decoder with several additional MLP layers, and lightweight token learning to enable object amsk and grasp map prediction. During training, the weights of the SAM modules are freezed, and only the adapter and the MLP layers in the decoder are updated. The learnable tokens consist of embedded token from prompts such as points or boxes obtained via a) mouse clicks, eye-gazing, or b) Grounding-DINO (G.D) and learnable tokens used to predict the object mask and grasp map.}

    \label{fig:graspsam}
\end{figure*}

\section{Method}

\subsection{Motivation}
With the growing adoption of visual foundation models (VFMs) in computer vision society, SAM (Segment Anything Model) has emerged as a leading solution, showcasing strong generalization in object segmentation. Building on this strength, we extend SAM to predict pixel-wise grasp quality maps for robotic manipulation. By leveraging SAM's robust pixel-wise classification abilities, its segmentation capabilities naturally transfer to grasp detection, enabling seamless integration of object identification and grasping within a single framework. This unified approach significantly improves the efficiency of robotic grasping tasks.

\subsection{Preliminaris of SAM.}
SAM is designed to segment any objects within an image using various types of prompt (i.e., points, boxes, or languages). The naive SAM is composed of three modules:
\begin{itemize}
    \item {\textbf{Image encoder:}} The image encoder is a ViT-based backbone designed to extract visual features from the input image.
    \item {\textbf{Prompt encoder:}} The prompt encoder transforms various input prompts into latent representations, providing positional information to help the model focus on regions of interest indicated by the prompts.
    \item {\textbf{Mask decoder:}} The mask decoder is a transformer-based module that uses features from the image encoder and combined tokens (learnable and prompt tokens from the prompt encoder) to predict the final mask.

\end{itemize}
While the original SAM model delivered impressive results, its large size limited practical use. To address this, lightweight version such as Mobile-SAM and Efficient-SAM were proposed, retaining SAM’s modular structure but optimizing for efficiency. GraspSAM builds on Efficient-SAM, tailored for grasp detection task.

\subsection{GraspSAM}
\noindent\textbf{GraspSAM Modules.}
To retain SAM’s zero-shot transfer capabilities while adapting it for grasp detection, we used a minimal adaptation approach. Rather than fully fine-tuning SAM or introducing a new decoder, we applied an adapter to the image encoder to enhance feature extraction for grasping. Additionally, we modified the existing SAM decoder by adding a few MLP layers to handle the prediction of refined object masks and grasp maps. Thus, GraspSAM consists of an image encoder with an adapter, a prompt encoder, and a combined mask and grasp decoder.

\noindent\textbf{Adapter for Grasp Detection.}
We adopted the Rein\cite{wei2024stronger} adapter to fine-tune the pretrained image encoder, enabling it to embed features specifically for object grasping while utilizing a minimal number of trainable parameters. As illustrated in Fig.\ref{fig:graspsam}, the $i$-th block $B_{i}$ of image encoder produce the features $f_{i}$ and the MLP-based adapter produces enhanced feature maps for the next block as follows.

\begin{equation}
    \begin{aligned}
        f_{1}&=B_{1}~(P.E(x))~~~~~~~~~~~~~~~~~~~~~~f_{1}\in \mathbb{R}^{n\times c}, \\ 
        f_{i+1}&=B_{i+1}(f_{i}+\hat{f_{i}})~~~~~~~~~~i=1,2,\ldots,N-1, \\ 
        f_{out}&=f_N+\hat{f_{N}},  
    \end{aligned}    
    \label{eq:delta}
\end{equation}

\noindent where $P.E(\cdot)$ denotes the patch embedding block in ViT-based image encoder, $n$ is the number of patches , $N$ represents the number of block, and $c$ is embedding dimension for the feature $f_1$. Note that parameters of encoder blocks $B_1, B_2,...,B_N$ are frozen, and only layers for adapter are training and generate features $\hat{f_{i}}$ as follows.

\begin{equation}
    \begin{aligned}
        \hat{f_{i}}&=Ad(f_i)~~~~~~f_{i}\in \mathbb{R}^{n\times \hat{c}}, i=1,2,\ldots,N-1,
    \end{aligned}    
    \label{eq:delta}
\end{equation}

\noindent where $Ad(\cdot)$ denotes the adapter, and $\hat{c}$ is each embedding dimension for the feature $f_i$.

\noindent\textbf{GraspSAM Output Tokens.}
Inspired by HQ-SAM’s high-quality mask prediction approach \cite{ke2024segment}, GraspSAM employs a similar token learning strategy\cite{carion2020end} to generate object masks and grasp maps tailored for robotic tasks. Previous works using SAM often rely on the pretrained model's mask outputs, which can lead to inaccuracies based on the prompt. To improve precision, GraspSAM introduces learnable mask tokens that adapt SAM’s mask predictions for grasp detection. Rather than fine-tuning the entire SAM model or adding a heavy decoder, we utilize learnable tokens for both mask and grasp predictions. These tokens are concatenated with SAM's original output and prompt tokens, creating a richer input for improved grasp detection. The mask and grasp tokens engage in self-attention and token-to-image mechanisms, enabling them to extract critical information from the image, prompt, and surrounding context for better accuracy. By training only these tokens and their associated layers, GraspSAM enhances SAM’s ability to predict both masks and grasps without altering its core architecture. This efficient approach preserves SAM’s zero-shot generalization while preventing overfitting, as only task-relevant components are updated.


\begin{figure*}[ht!]
    \centering
        \includegraphics[width=0.8\textwidth]{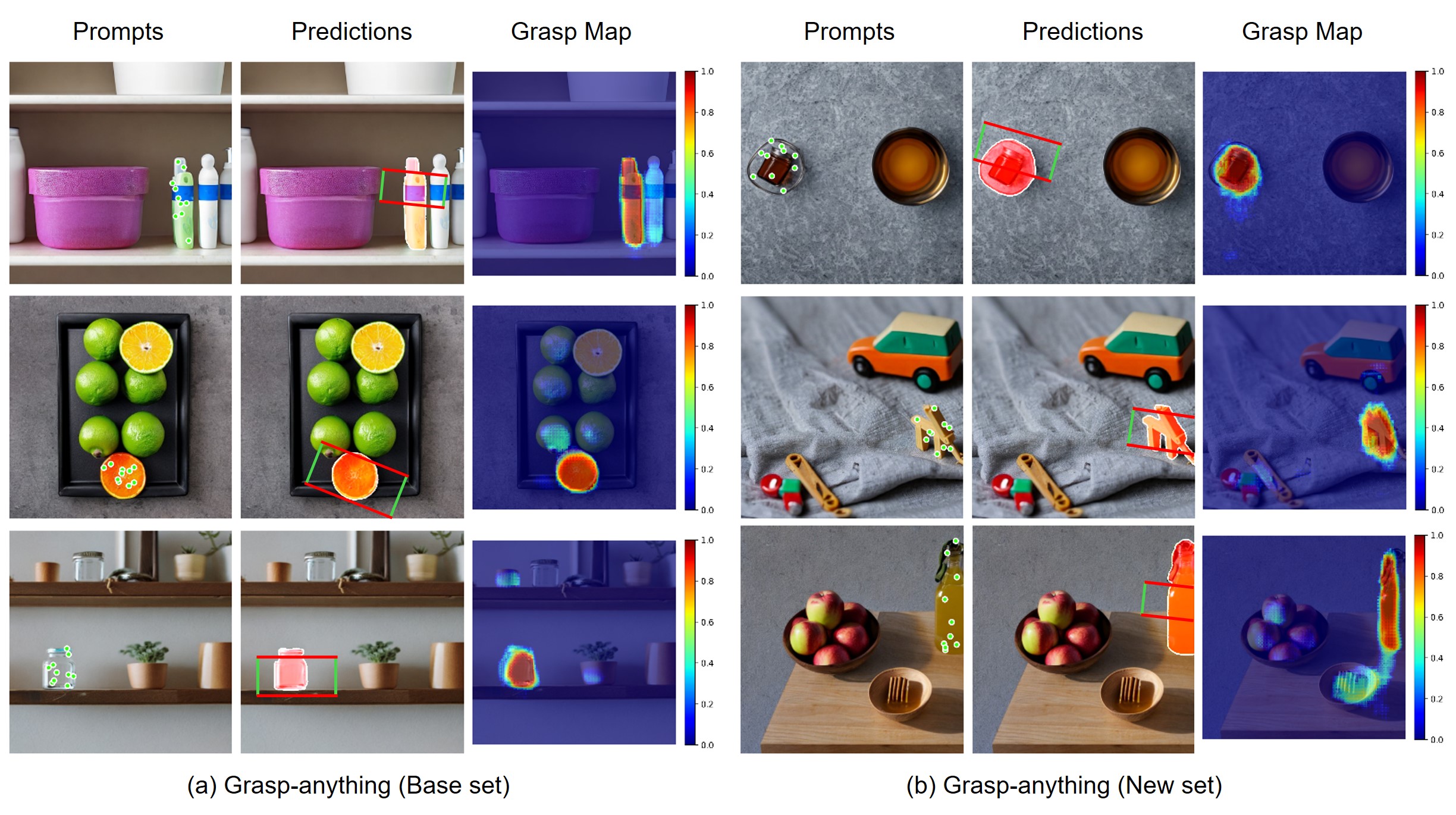}
    \caption{\textbf{Visualization of GraspSAM prediction.} We visualize the input prompts (10 points) along with the predicted outputs, including the object mask and grasp box. Additionally, we display the predicted grasp quality map. (a) corresponds to the Grasp-Anything Base set, while (b) represents the New set.}
    \label{fig:predictions}
\end{figure*}

\noindent\textbf{GraspSAM Decoder.}
The Mask and Grasp Decoder in GraspSAM combines SAM's original decoder with additional components to produce refined object masks and grasp maps. We designed the G.S. Block (Grasp-SAM Block) in Fig.\ref{fig:graspsam} to fuse multi-scale features from the image encoder, enhancing both global context and local detail representation. Following the G.S. Blocks, we attach MLP layers and grasp heads, which include a grasp confidence head, gripper angle head, gripper width head, and object mask head. These components output heatmaps for grasp confidence, gripper angle, gripper width, and refined object masks. This design allows GraspSAM to efficiently generate precise grasp predictions and object masks with a single forward pass, maintaining both accuracy and computational efficiency.

\noindent\textbf{Loss Functions.}
Our loss function comprise two terms; object mask loss as $\mathcal{L}_{mask}$ and grasp detection loss $\mathcal{L}_{grasp}$,

\begin{equation}
\label{eq:the_sum_of_losses}
\mathcal{L} = \lambda_1 * \mathcal{L}_{mask} + \lambda_2 * \mathcal{L}_{grasp},
\end{equation}

\noindent where $\lambda_1$ and $\lambda_2$ are hyper-parameters, setting as $\lambda_1=2$ and $\lambda_2=1$ respectively. To encourage the accurate prediction of the object mask for the specified by the prompt (e.g., points, box) among multiple objects in the grasp dataset, we employed MSE loss for the object mask loss $\mathcal{L}_{mask}$. We developed an object grasping loss $\mathcal{L}_{grasp}$ to train grasp detection for the target object. This grasping loss applies a higher weight to the foreground and a lower weight to the background, based on the ground-truth object mask, to ensure that the grasp heatmap is learned effectively for the target object. The object grasping loss formulated as follows,

\begin{equation}
\label{eq:the_grasping_loss}
\mathcal{L}_{grasp} = \lambda_3 * \mathcal{L}_{fore} + \lambda_4 * \mathcal{L}_{back},
\end{equation}

\noindent where $\lambda_3$ and $\lambda_4$ are set as 1 and 0.01 respectively. 

\noindent\textbf{Training Details.}
We trained GraspSAM on the Jacquard and Grasp-Anything datasets with a batch size of 8, using 4 NVIDIA RTX 3090 GPUs for 50 epochs. We did not apply data augmentation and train the model with a learning rate of 1e-5, using the AdamW optimizer and cosine annealing learning rate scheduler.

\section{Experiments}
\noindent\textbf{Datasets.} 
To ensure a fair comparison, we followed the experimental settings of the GG-CNN\cite{morrison2018closing} and LGD\cite{vuong2024language}. We trained and evaluated GraspSAM and the baseline models on the grasp benchmark datasets, Jacquard\cite{depierre2018jacquard}, Grasp-Anything\cite{vuong2023grasp} and Grasp-Anything++\cite{vuong2024language} datasets. We utilized the Base and New sets as defined in the Grasp-Anything\cite{vuong2023grasp}. The Base set includes the top 70\% most frequent labels from the LVIS dataset\cite{gupta2019lvis}, while the New set consists of the remaining 30\% less frequent labels.

\begin{table}[ht]
\centering
\caption{Grasp detection performance of each model given 10 points as prompt. The $*$ symbol indicates that Efficient-SAM (ViT-t) performs object masking using the prompt, and the following grasp detection models predict the grasp for the masked object. GraspSAM-tiny and GraspSAM-t refer to using MobileSAM (Tiny-ViT) and Efficient-SAM (ViT-t) as backbones, respectively. \textbf{Bold} and \underline{underline} mean the best result and second best result respectively.}
\label{tab:main_table}
\resizebox{0.95\columnwidth}{!}{%
\begin{tabular}{c|ccc|ccc}
\hline
\multirow{2}{*}{Methods}      & \multicolumn{3}{c|}{Grasp-Anything\cite{vuong2023grasp}} & \multicolumn{3}{c}{Jacquard\cite{depierre2018jacquard}} \\ \cline{2-7}
          & Base      & New         & H         & Base    & New    & H      \\ \hline
GR-ConvNet$^*$\cite{kumra2020antipodal}       & 0.68      & 0.55        & 0.61      & 0.82    & 0.61      & 0.70   \\ \hline
Det-Seg-Refine$^*$\cite{ainetter2021end}   & 0.58      & 0.53        & 0.55      & 0.79    & 0.55       & 0.65   \\ \hline
GG-CNN$^*$\cite{morrison2018closing}           & 0.65      & 0.53        & 0.58      & 0.73    & 0.52      & 0.61   \\ \hline
LGD$^*$\cite{vuong2024language}              & 0.69      & 0.57        & 0.62       & 0.83    & 0.64      & 0.72   \\ \hline \hline

GraspSAM-tiny (ours)  & \underline{0.78}   & \underline{0.75}  & \underline{0.77}  & \textbf{0.90}  & \textbf{0.81} & \textbf{0.85}   \\ \hline
GraspSAM-t (ours)  & \textbf{0.83}   & \textbf{0.81}  & \textbf{0.82}  & \underline{0.87}  & \underline{0.75} & \underline{0.81}   \\ \hline
\end{tabular}%
}
\end{table}


\begin{table}[ht]
\footnotesize\addtolength{\tabcolsep}{+8.24pt}
\caption{Grasp dection performance comparison when using language as a prompt. "G.D" refers to Grounding-Dino.}
\label{tab:language-driven-grasp-detection}
\begin{tabular*}{\columnwidth}{cccc}
\hline
\multirow{2}{*}{Methods}     & \multicolumn{3}{c}{Grasp-anthing ++\cite{vuong2024language}} \\ \cline{2-4} 
                             & Base         & New        & H        \\ \hline
\multicolumn{1}{c|}{CLIPORT\cite{shridhar2022cliport}} &           0.36 &                0.26 &  0.29         \\
\multicolumn{1}{c|}{CLIPGrasp\cite{xu2023joint}} &          0.40 &                0.29 &  0.33         \\
\multicolumn{1}{c|}{LGD\cite{vuong2024language}}     &           0.48 &                0.42 &  0.45         \\
\multicolumn{1}{c|}{GraspSAM w/ G.D (Ours)} &  \textbf{0.64} &       \textbf{0.62} &  \textbf{0.63}     \\ \hline
\end{tabular*}%
\end{table}


\noindent\textbf{Baselines.}
We set up GraspSAM using Efficient-SAM (ES)\cite{xiong2024efficientsam} as the backbone, trained with 10 points as prompts. For comparison, we evaluated other grasp detection methods, including GR-ConvNet\cite{kumra2020antipodal}, Det-Seg-Refine\cite{ainetter2021end}, GG-CNN\cite{morrison2018closing} and LGD (no text version)\cite{vuong2024language}. Since these baseline models do not accept prompts as input, we used a pre-trained ES model for object identification. The output masks from ES were then used by the grasp detection models to predict grasps for the identified object.

\noindent\textbf{Metrics.}
Our primary metric is the success rate, defined in line with prior works \cite{kumra2020antipodal, morrison2018closing}. A predicted grasp is considered successful if it achieves an Intersection over Union (IoU) score greater than 25\% with the ground truth grasp and has an offset angle of less than $30^\circ$. Additionally, if the mask predicted a different object than the one specified by the prompt, it was considered a failure. To measure overall performance across different categories, we employ the harmonic mean (‘H’) of success rates \cite{zhou2022conditional}, which allows for a comprehensive assessment of GraspSAM's generalization ability.



\begin{table}[ht]
\footnotesize\addtolength{\tabcolsep}{+6.5pt}
\caption{Cross-dataset grasp detection results (Left: GR-ConvNet\cite{kumra2020antipodal}, Right: GraspSAM (Ours)).}
\label{tab:cross-data-val}
\begin{tabular*}{\columnwidth}{c|cc}
\hline
Train\textbackslash{}Test & Grasp-Anything\cite{vuong2023grasp} & Jacquard\cite{depierre2018jacquard}  \\ \hline
Grasp-Anything & 0.68 / \textbf{0.83} & 0.37 / \textbf{0.62}        \\ 
Jacquard       & 0.16 / \textbf{0.27} & 0.82 / \textbf{0.87} \\ \hline
\end{tabular*}%
\end{table}



\begin{table}[ht]
\footnotesize\addtolength{\tabcolsep}{+0.5pt}
\caption{Performance of GraspSAM with and without Adapter.}
\label{tab:comparision of adapter}
\begin{tabular*}{\columnwidth}{c|ccc|ccc}
\hline
\multirow{2}{*}{Methods}    & \multicolumn{3}{c|}{Grasp-Anything\cite{vuong2023grasp}} & \multicolumn{3}{c}{Jacquard\cite{depierre2018jacquard}} \\ \cline{2-7} 
                     & Base       & New       & H       & Base     & New     & H    \\ \hline
GraspSAM w/o AD &        0.80&          0.75&     0.77&      0.86&        0.66&     0.75   \\ \hline
GraspSAM w/ AD  &        \textbf{0.83}& \textbf{0.81}& \textbf{0.82}&  \textbf{0.87}& \textbf{0.75}& \textbf{0.81}   \\ \hline
\end{tabular*}%
\end{table}



\begin{table}[ht]
\footnotesize\addtolength{\tabcolsep}{+1.2pt}
\caption{Performance of GraspSAM (GS) across different adapters.}
\label{tab:comparision of adapter type}

\begin{tabular*}{\columnwidth}{c|ccc|ccc}
\hline
\multirow{2}{*}{Methods}    & \multicolumn{3}{c|}{Grasp-Anything\cite{vuong2023grasp}} & \multicolumn{3}{c}{Jacquard\cite{depierre2018jacquard}} \\ \cline{2-7} 
                     & Base       & New       & H       & Base     & New     & H    \\ \hline
GS + LoRA\cite{hu2021lora}&     0.81       &   0.77       &    0.79  &   \textbf{0.87}   &  0.69      &  0.77       \\ \hline
GS + Rein\cite{wei2024stronger}&   \textbf{0.83}& \textbf{0.81}& \textbf{0.82}&  \textbf{0.87}& \textbf{0.75}& \textbf{0.81}   \\ \hline
\end{tabular*}%
\end{table}

\begin{figure}[ht!]
    \centering
        \begin{subfigure}[t]{\columnwidth}
            \centering
            \includegraphics[width=0.8\textwidth]{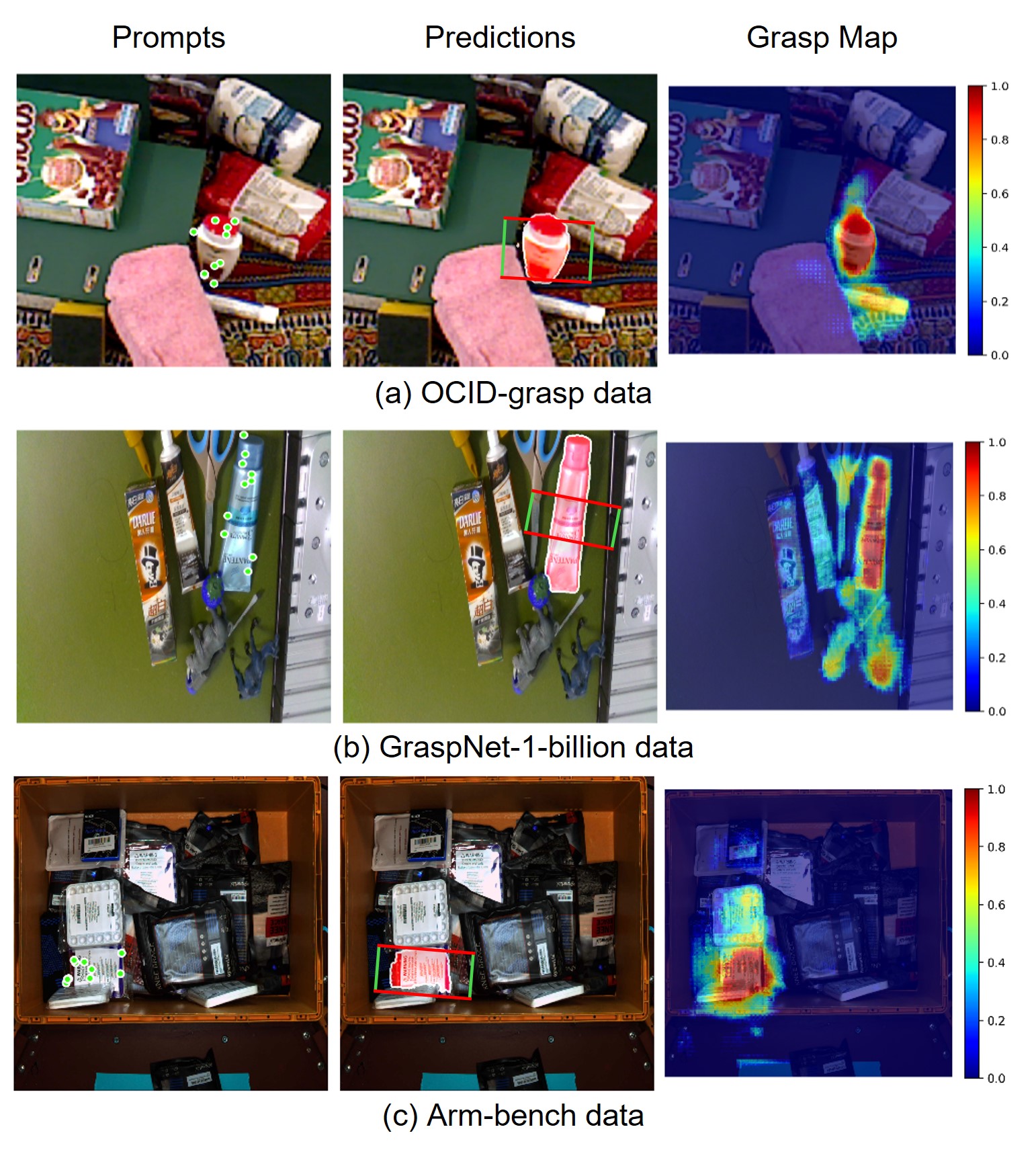}
        \end{subfigure}
        
    \caption{Visualization of in-the-wild grasp detection results}
    \label{fig:inthewild}
\end{figure}

\noindent\textbf{Prompt-driven Grasp Detection.} GraspSAM was evaluated using MobileSAM (Tiny-ViT) and EfficientSAM (ViT-t) backbones. As shown in Table \ref{tab:main_table}, GraspSAM achieved state-of-the-art success rates on the Grasp-Anything and Jacquard benchmarks. Further analysis of the New set and H metric shows that GraspSAM retains SAM’s zero-shot segmentation ability while effectively learning grasp prediction. GraspSAM also supports language prompts for grasp detection. Trained and evaluated on Grasp-Anything++ using the same settings as LGD, we used Grounding-DINO to convert language to bounding box, which were then used as prompts for GraspSAM. As shown in Table \ref{tab:language-driven-grasp-detection}, GraspSAM outperformed the LGD model, demonstrating its versatility in handling different prompt types.

\noindent\textbf{Category-agnostic Grasp Detection.}
We compared GraspSAM's category-agnostic grasp detection with existing methods using Grounding-Dino and the fixed prompt "A rigid object."\cite{fang2024embodied}. All generated bounding boxes were used as prompts, and failure was defined if no bounding boxes were produced. While GraspSAM did not achieve the best performance on the Jacquard Base set, it significantly outperformed other methods on the remaining sets, demonstrating its robustness even without precise object prompts (Table\ref{tab:ca_grasp_detection}).

\begin{table}[ht]
\centering
\caption{Category-agnostic grasp detection performance. \textbf{Bold} and \underline{underline} mean the best result and second best result respectively.}
\label{tab:ca_grasp_detection}
\resizebox{0.95\columnwidth}{!}{%
\begin{tabular}{c|ccc|ccc}
\hline
\multirow{2}{*}{Methods}     & \multicolumn{3}{c|}{Grasp-Anything\cite{vuong2023grasp}} & \multicolumn{3}{c}{Jacquard\cite{depierre2018jacquard}} \\ \cline{2-7} 
                 & Base      & New         & H         & Base    & New    & H      \\ \hline
GR-ConvNet\cite{kumra2020antipodal}       & 0.75      & 0.61        & 0.67      & \underline{0.88}    & 0.66      & 0.75   \\ \hline
Det-Seg-Refine\cite{ainetter2021end}   & 0.64      & 0.59        & 0.61      & 0.85    & 0.59       & 0.70   \\ \hline
GG-CNN\cite{morrison2018closing}           & 0.72      & 0.59        & 0.65      & 0.78    & 0.56      & 0.65   \\ \hline
LGD\cite{vuong2024language}              & 0.77      & 0.65        & 0.70      & \textbf{0.89}  & 0.70      & 0.78   \\ \hline \hline

GraspSAM-tiny (ours)  & \underline{0.79} & \underline{0.68} & \underline{0.73}      & \underline{0.88}    & \textbf{0.79}      & \textbf{0.83} \\ \hline
GraspSAM-t (ours)     & \textbf{0.89} & \textbf{0.82}  & \textbf{0.85}      & 0.83    & \underline{0.72}      & \underline{0.77} \\ \hline
\end{tabular}%
}
\end{table}

\begin{table}[ht]
\footnotesize\addtolength{\tabcolsep}{15.4pt}
\caption{GraspSAM results for different prompt types.}
\label{tab:cross-prompt-eval}
\begin{tabular*}{\columnwidth}{cccc}
\hline
\multirow{2}{*}{Prompt}     & \multicolumn{3}{c}{Grasp-anthing\cite{vuong2023grasp}} \\ \cline{2-4} 
                             & Base         & New        & H        \\ \hline
\multicolumn{1}{c|}{1 point}     &  0.78 &       0.73 &  0.75     \\
\multicolumn{1}{c|}{3 points}     & 0.83 &       0.80 &  0.81     \\
\multicolumn{1}{c|}{5 points}     & 0.83 &       0.80 &  0.81     \\
\multicolumn{1}{c|}{10 points}   &  0.83 &       0.81 &  0.81     \\ 
\multicolumn{1}{c|}{Box}         &  \textbf{0.85} &      \textbf{0.82} &  \textbf{0.82}     \\ 
\hline
\end{tabular*}%
\end{table}


\noindent\textbf{Cross-dataset Grasp Detection.}
We conducted cross-dataset validation to assess the zero-shot performance of GraspSAM across different data domains. We compared GraspSAM's cross-dataset validation performance with the CNN-based state-of-the-art model, GR-ConvNet\cite{kumra2020antipodal}. While GR-ConvNet showed significant performance drops when transitioning between different data domains, GraspSAM exhibited relatively minor declines, demonstrating its robustness and superior generalization capabilities (Table \ref{tab:cross-data-val}).

\noindent\textbf{In-the-wild Grasp Detection.}
Figure \ref{fig:inthewild} visualizes the predictions of GraspSAM, trained on the Grasp-Anything dataset, across various real-world datasets (OCID-grasp\cite{ainetter2021end}, GraspNet\cite{fang2020graspnet}, Armbench\cite{mitash2023armbench}) reflecting domestic or industrial environments. The results demonstrate GraspSAM’s robustness in handling complex background textures (Fig. \ref{fig:inthewild}-(a)), heavily cluttered objects (Fig. \ref{fig:inthewild}-(b)), and when prompted to grasp occluded objects (Fig. \ref{fig:inthewild}-(c)).

\noindent\textbf{Effectiveness of Adapter.}
To enable SAM to learn features for grasp detection efficiently, we employed adapters. To validate this, we compared a model with the SAM encoder frozen and the decoder trained without adapters. As shown in Table\ref{tab:comparision of adapter}, models trained with adapters performed better, especially on the New set, demonstrating their role in improving generalization and accuracy. We further compared different adapter types, finding that the Rein adapter outperformed the widely-used LoRA\cite{hu2021lora} adapter, particularly on the Grasp-Anything dataset's New set, showing that Rein\cite{wei2024stronger} maintains SAM’s generalization while enhancing grasp detection (Table\ref{tab:comparision of adapter type}).

\begin{figure}[ht!]
    \centering
        \begin{subfigure}[t]{\columnwidth}
            \centering
            \includegraphics[width=\textwidth]{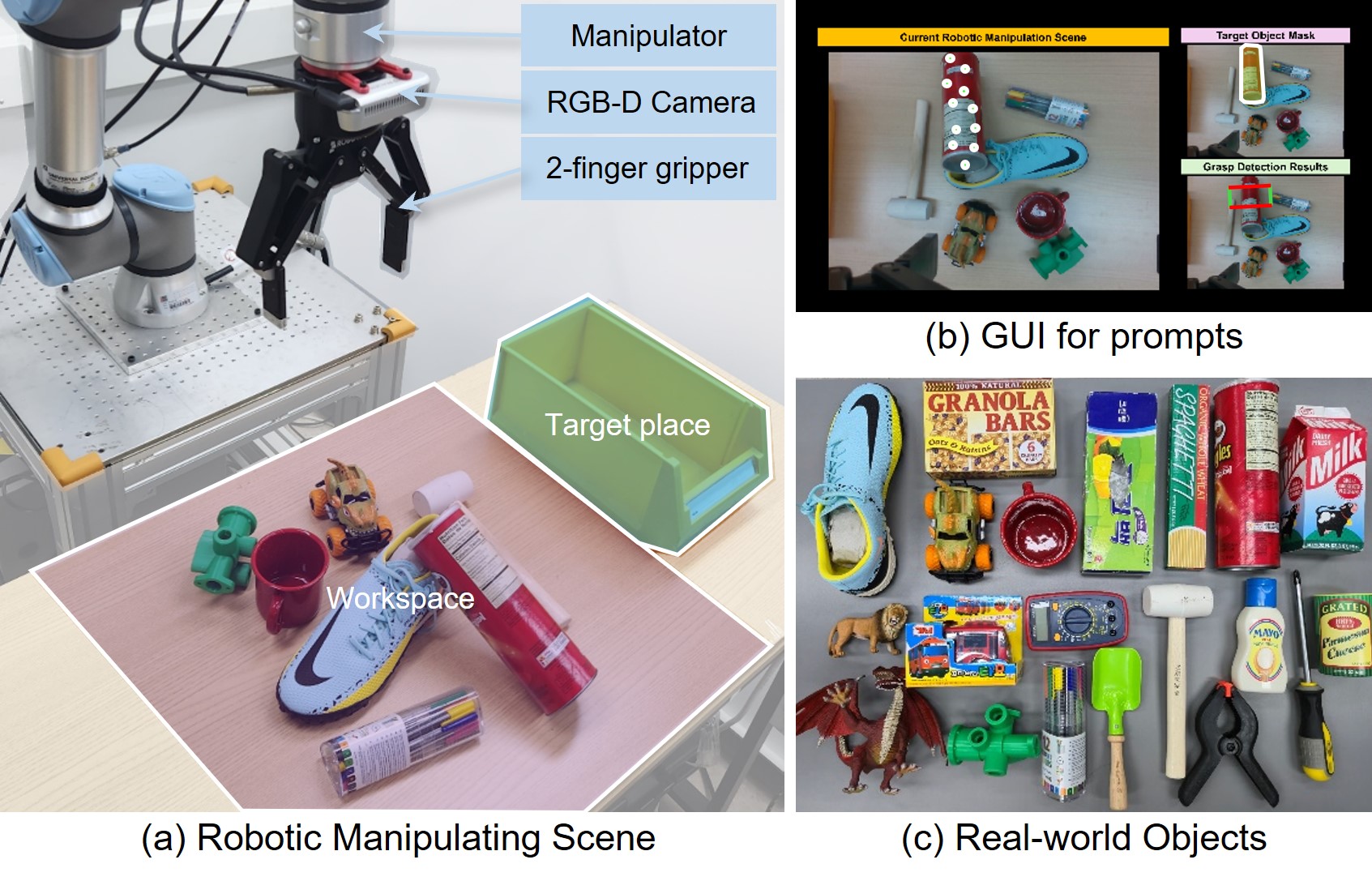}
        \end{subfigure}
        
    \caption{Real-world experiments settings.}
    \label{fig:realworldsetting}
\end{figure}


\begin{figure}[ht]
    \centering
        \begin{subfigure}[t]{\columnwidth}
            \centering
            \includegraphics[width=\textwidth]{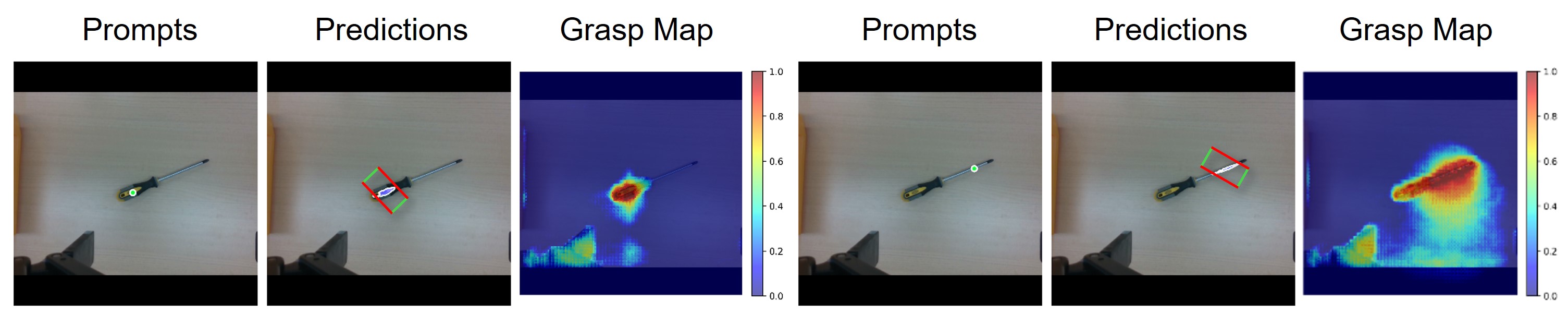}
        \end{subfigure}
        
    \caption{Grasp detection results based on prompt location.}
    \label{fig:affordance_grasp}
\end{figure}


\noindent\textbf{Additional Experiments.}
We conducted experiments to compare GraspSAM’s grasp detection performance based on different prompt types (1 point, 3 points, 5 points, 10 points, and box) and backbone configurations. As shown in Table \ref{tab:cross-prompt-eval}, using Box prompts for both training and evaluation on the GA dataset yielded the best performance, while providing a single point as the prompt resulted in the lowest performance, though it still outperformed the baseline methods listed in Table \ref{tab:main_table}. Starting from 3-point prompts, we observed significant improvements in grasp detection accuracy, with diminishing returns in performance gains as the number of points increased. Overall, the best results were obtained with box prompts.

\begin{table}[ht]
\centering
\caption{Grasp detection performance and inference comparison based on different backbone types.}
\label{tab:grasp_det_eval_for_various_backbones}
\resizebox{\columnwidth}{!}{%
\begin{tabular}{ccccccc}
\hline
\multirow{2}{*}{Backbones}    & \multicolumn{3}{c}{Grasp-Anything\cite{vuong2023grasp}} & \multirow{2}{*}{Params (M) $\downarrow$} & \multirow{2}{*}{\begin{tabular}[c]{@{}l@{}}Trainable \\ Params (M)$\downarrow$\end{tabular}} & \multirow{2}{*}{FLOPs(G)$\downarrow$} \\ \cline{2-4}
  & Base  & New   & H    &     &   &                                       \\ \hline
\multicolumn{1}{c|}{Mobile-sam (Tiny-ViT)} & 0.78  & 0.75  & 0.77   & 15.26     & 1.12 & 52.03 \\
\multicolumn{1}{c|}{Efficient-SAM (ViT-t)} & 0.83  & 0.81  & 0.82   & 15.39     & 1.15 & 114.96 \\
\multicolumn{1}{c|}{Efficient-SAM (ViT-s)} & \textbf{0.85} & \textbf{0.82} & \textbf{0.83} & 32.39 & 1.78 & 268.80 \\ \hline
\end{tabular}%
}
\end{table}

\begin{table}[]
\footnotesize\addtolength{\tabcolsep}{+7.5pt}
\caption{Grasp performance in the real-world.}
\label{tab:real-world-exp}
\begin{tabular*}{\columnwidth}{c|cc}
\hline
Methods           & Physical grasp & Success rate (\%) \\ \hline
GG-CNN$^*$\cite{morrison2018closing} &     68 / 100   &     68         \\
GraspSAM (Ours)   &     \textbf{86 / 100}   &     \textbf{86}         \\ \hline
\end{tabular*}%
\end{table}

Additionally, Table \ref{tab:grasp_det_eval_for_various_backbones} reports the results for three backbone types (Tiny-ViT, ViT-t, ViT-s), indicating that grasp detection performance improves proportionally to the number of parameters, although inference time also increases, presenting a trade-off. We also measured the trainable parameters of each GraspSAM variant, which include only the adapter and modified decoder. Notably, even with just 1/10 of the total model’s parameters, GraspSAM effectively learns grasp detection, balancing efficiency with performance.

\section{Prompt-driven grasping in the real-world}
As shown in Figure \ref{fig:realworldsetting}-(a), we conducted the grasp experiment using a UR5e robot, a Robotiq 2f-140 gripper, and a RealSense D435 RGB-D camera. We selected 20 household or industrial objects and placed 6 objects in a cluttered arrangement per scenario (Figure \ref{fig:realworldsetting}-(c). GraspSAM predicted object masks and grasp poses using 10 point prompts from randomly GUI clicks on the target object, combined with the robot’s view (Figure \ref{fig:realworldsetting}-(b). The robot then executed the grasp with a motion planner. A grasp was considered a failure if the wrong object was targeted or if the robot failed to lift it by 20 cm. We evaluated 100 scenarios, attempting 5 grasps per object across 20 objects. GraspSAM achieved an 86\% success rate, outperforming the two-stage method with EfficientSAM and GGCNN (Table \ref{tab:real-world-exp}). We also visualized GraspSAM’s task-oriented grasp capabilities with only 1 point as prompt. In Figure \ref{fig:affordance_grasp}-(a), the prompt on a screwdriver's bit led to a grasp on the bit, while a prompt on the handle (Figure \ref{fig:affordance_grasp}-(b)) focused on the handle, showcasing GraspSAM’s potential for task-specific grasping.

\section{Conclusion}
We presented GraspSAM, an extension of SAM for end-to-end grasp detection that unifies object segmentation and grasp planning into a single framework. By introducing adaptation methods and lightweight modifications to the decoder, GraspSAM retains SAM’s generalization abilities while efficiently learning grasp prediction. Extensive evaluations showed state-of-the-art (SOTA) performance in category-agnostic and prompt-driven tasks across the Jacquard and Grasp-Anything datasets, as well as robust real-world applicability. GraspSAM’s ability to handle diverse prompt types highlights its versatility in practical settings. Future work will focus on expanding GraspSAM's grasp capabilities to 6-DOF and incorporating a dedicated language encoder for direct, end-to-end language-driven grasp detection.



\section*{Acknowledgment}
\begin{spacing}{0.4}
{\scriptsize This work was supported by Institute of Information \& communications Technology Planning \& Evaluation (IITP) grant funded by the Korea government(MSIT) No.RS-2021-II212068, Artificial Intelligence Innovation Hub.}
\end{spacing}

\bibliography{bibtex/bib/references.bib}{}
\bibliographystyle{IEEEtran}

\end{document}